\documentclass{article}
\usepackage{spconf,amsmath,graphicx}

\usepackage{amssymb}
\usepackage{booktabs}
\usepackage{multirow}
\usepackage{cite}
\usepackage{subfig}

\usepackage[capitalize]{cleveref}
\crefname{section}{Sec.}{Secs.}
\Crefname{section}{Section}{Sections}
\Crefname{table}{Table}{Tables}
\crefname{table}{Tab.}{Tabs.}

\title{LLA-FLOW: A Lightweight Local Aggregation on Cost Volume for Optical Flow Estimation}
\name{Jiawei Xu\qquad  Zongqing Lu$^{*}$\qquad  
Qingmin Liao$^{*}$\thanks{$^{*}$Corresponding author.}}
\address{Shenzhen International Graduate School, Tsinghua University, China}
\begin{document}
%
\maketitle
\begin{abstract}
Lack of texture often causes ambiguity in matching, and handling this issue is an important challenge in optical flow estimation. Some methods insert stacked transformer modules that allow the network to use global information of cost volume for estimation. But the global information aggregation often incurs serious memory and time costs during training and inference, which hinders model deployment. We draw inspiration from the traditional local region constraint and design the local similarity aggregation (LSA) and the shifted local similarity aggregation (SLSA). The aggregation for cost volume is implemented with lightweight modules that act on the feature maps. Experiments on the final pass of Sintel show the lower cost required for our approach while maintaining competitive performance.
\end{abstract}
\begin{keywords}
Optical flow, Neural networks, Attention mechanism, Local aggregation
\end{keywords}

\section{Introduction}
\label{sec:intro}
The goal of the optical flow estimation is to estimate the 2D motion information of each pixel using a sequence of frames which can be used as a base input for many high-level tasks, such as video super resolution\cite{liao2015video}, video interpolation\cite{splat}, video segmentation\cite{seg} and pose estimation\cite{pose}.
With the development of deep learning techniques, the field of optical flow estimation has advanced significantly.

Cost volume is a fundamental component of deep learning methods for optical flow estimation. The earliest optical flow neural network, FlowNet\cite{flownet}, demonstrated its importance through experiments. Cost volume is calculated by the dot product operation between the features of two frames and represents the correlation value or confidence level between pixels.
PWC-Net\cite{pwc} computes a partial cost volume with a limited range. It is a typical approach that applies the coarse-to-fine strategy, but the strategy struggles with the problem of losing small fast-moving objects at low resolution.
RAFT\cite{raft} introduces all-pairs field transforms to construct a complete pixel-to-pixel matching relationship known as 4D cost volume.
Considering the enormous size of cost volume, RAFT introduces a multi-scale cost volumes strategy and index operation, which uses partial correlation values to construct correlation features for subsequent calculation rather than the original 4D cost volume, effectively addressing the large displacements challenge.
However, due to other challenges that typically arise in the optical flow estimation, such as lack of texture, the image features are not representative of the original scene and the 4D cost volume contains many unreliable correlation values or outliers.
If these outliers are selected by the index operation, they will lead to a misunderstanding of motion in subsequent modules.

An effective solution is to directly smooth the cost volume.
Separable Flow\cite{sflow} designs the separable cost volume computation module, which compresses the 4D cost volume to two 3D cost volumes along the horizontal and vertical orientations. Then the approach uses stacked non-local aggregation layers to refine the 3D cost volumes separately.
They are concatenated and fed into the update operator.
FlowFormer\cite{flowformer} abandons the multi-scale cost volumes strategy and builds encoder and decoder modules by stacking transformers for cost volume.
It first uses a convolutional neural network to patchify the 4D cost volume, then projects patches into latent cost tokens.
To generate the cost memory, the approach proposes alternate-group transformer layers
that aggregate tokens within the same cost map and tokens from different cost maps. 
A recurrent transformer decoder with dynamic positional cost queries is used for decoding.
The impact of outliers in the cost volume is reduced by aggregation.
However, these approaches have to compress or encode the cost volume and stack repeated modules to maintain reliability, which causes severe time overhead.

In traditional algorithms\cite{lk}, it is assumed that if neighboring pixels are similar, their optical flow results are also similar.
Based on this constraint assumption, we design the local aggregation modules for the 4D cost volume to suppress the outliers.
We allow each correlation value to be influenced by its neighbors in the same cost map with weights determined according to similarity.
The cost maps of pixels in the local region are shifted and summed so that they can use the matching information of neighbors to modify their cost map.
The aggregated cost volume shows more clear and more proper matching information than before.
Based on the 4D cost volume generation formula, we present lightweight operations that act on the feature maps of both frames to perform the aggregation.

Our contributions are summarized as follows: 
(1) We design the local aggregation for 4D cost volume and present lightweight operations to diminish the impact of outliers caused by lack of texture.
(2) We thoroughly compared the local aggregation for cost volume with published global aggregation methods in terms of parameter, training memory, inference memory, and runtime.
(3) We apply our modules on RAFT at a low cost and demonstrate competitive performance on the final pass of Sintel.

\section{Methodology}
\label{sec:Methodology}
The cost volume provides clues for the optical flow estimation, and the higher the correlation value, the more likely the two pixels will be paired.
We allow the correlation values of the pixel pairs to be influenced by their neighbors.
It is challenging for traditional local aggregation methods such as convolutional layers\cite{cnn} to adapt the parameters to varied contexts.
Non-local operations like transformer\cite{transformer}, which utilize point similarity to calculate point weights, are more adaptable to different contexts. 
Our method consider only a limited range of surrounding points and determine their weights based on similarity score.
The first two dimensions and the last two dimensions of the 4D cost volume are aggregated separately, to represent the aggregation between cost maps and the aggregation inside a single cost map.

First, we use the feature extractor to obtain the features $F_1,F_2 \in \mathbb{R}^{H\times W\times C}$, and then apply dot product operation to calculate the 4D cost volume $C\in \mathbb{R}^{H\times W\times H\times W}$,

\begin{figure}[htb]
  \subfloat[The local similarity aggregation (LSA).]{\includegraphics[width=1\linewidth]{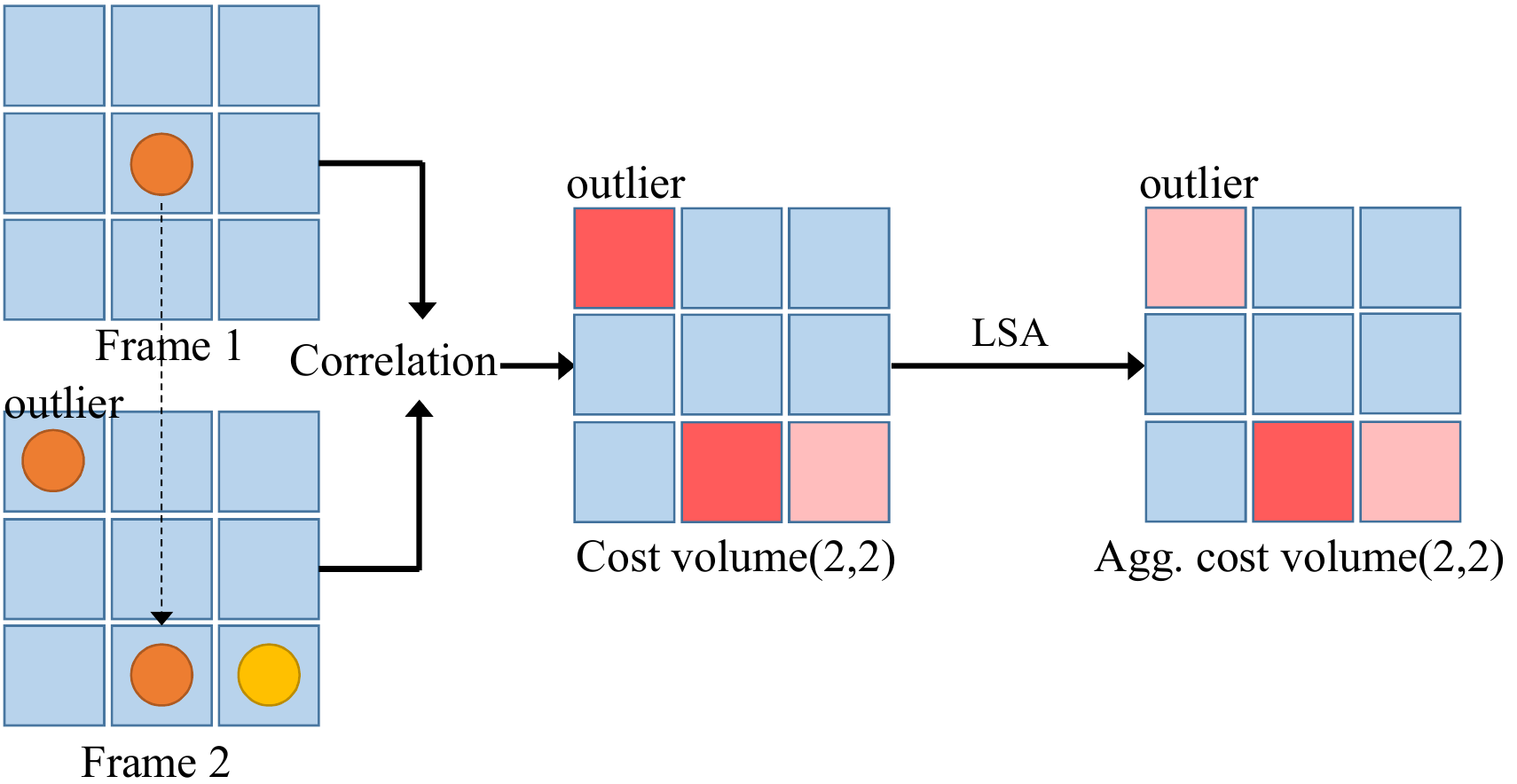}\label{fig:LSA}}
  \newline 
  \subfloat[The shifted local similarity aggregation (SLSA).]{\includegraphics[width=1\linewidth]{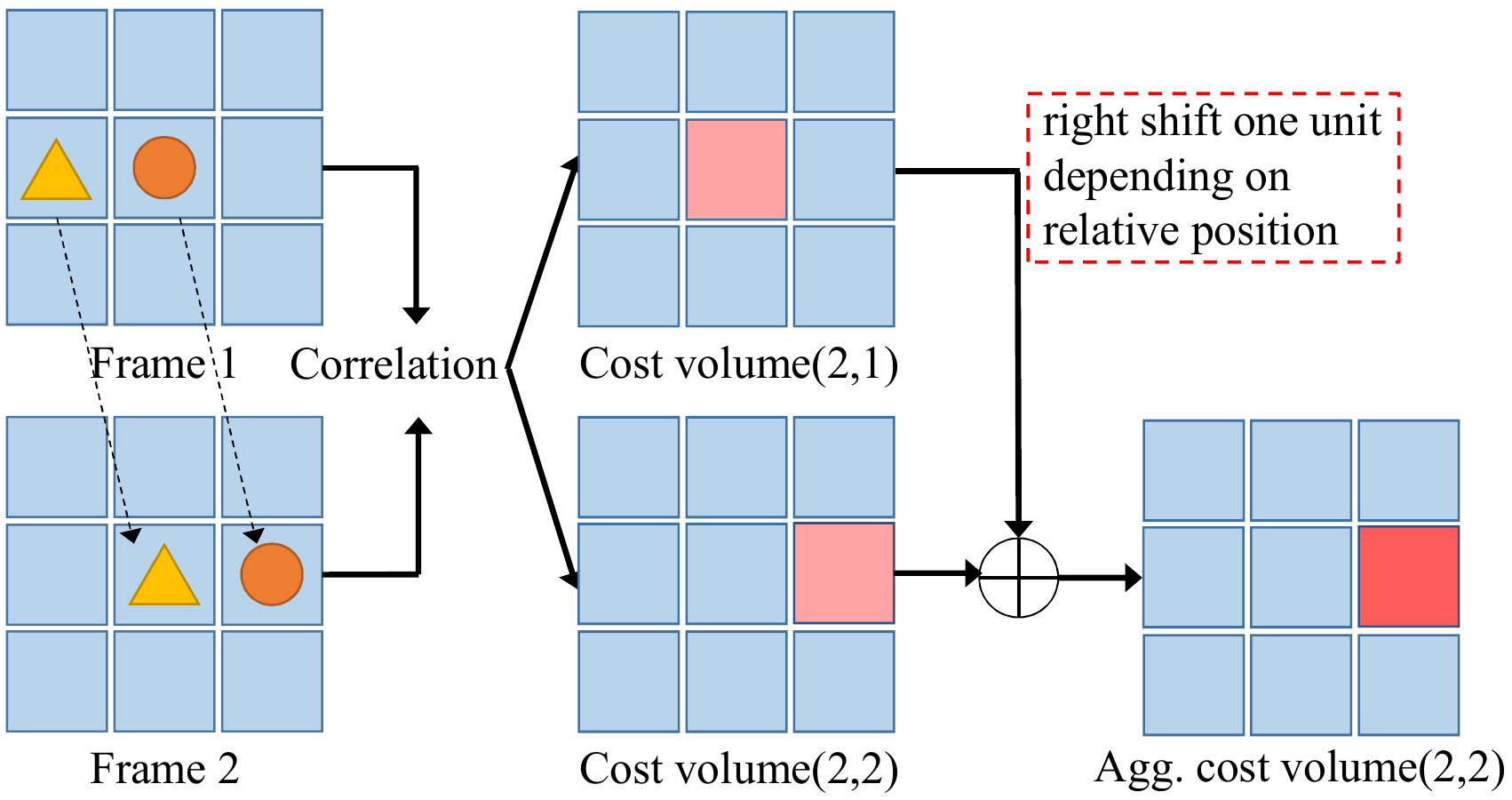}\label{fig:SLSA}}
\caption{Visualization of the aggregation. Dark colours represent high correlation values
.}
\label{fig:agg}
\end{figure}

\begin{figure*}[htb]
  \centering
  \includegraphics[width=1\linewidth]{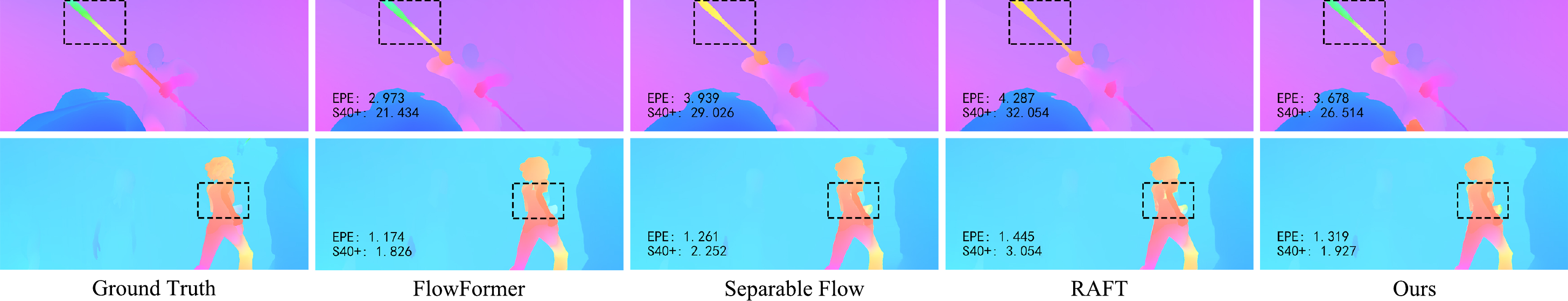}
\caption{Visual results on the Sintel test set.}
\label{fig:test}
\end{figure*}

\begin{equation}
\begin{aligned}
C(i,j,m,n)
&=\sum\limits_c F_1(i,j,c)\cdot F_2(m,n,c)\\
&=\big\langle F_1(i,j),F_2(m,n) \big\rangle
\end{aligned}
\label{eq:fomula}
\end{equation}
Based on \cref{eq:fomula}, we present lightweight operations to perform the aggregation. More details are shown below.

\subsection{Local Similarity Aggregation}
If the aggregation is performed within a single cost map, the aggregated cost volume can be expressed as,
\begin{equation}
\begin{aligned}
C'(i,j,m,n)
&=\sum_{(m_k,n_k)\in lr}\mathcal{W}_{lsa}\cdot C(i,j,m_k,n_k)\\
&=\big\langle F_1(i,j),\sum_{(m_k,n_k)}^{lr}\mathcal{W}_{lsa}\cdot F_2(m_k,n_k) \big\rangle
\end{aligned}
\label{eq:lsa}
\end{equation}
where $lr$ represents the local region and $(m_k,n_k)$ are the coordinates in the features $F_2$. The operator apply on the tensor $F_2$, which is much smaller than the cost volume, since the \cref{eq:lsa} shows that the aggregation for the cost volume is identical to the aggregation for $F_2$.
As mentioned before, 
the weights are determined by the similarity scores between the points.
We use the context features $F_{c}$ obtained by the context encoder to calculate the weights $\mathcal{W}_{lsa}$ rather than the features $F_2$. We use $x$ to denote the center point $F_{c}(m,n)$, $x_k$ to denote the weight point $F_{c}(m_k,n_k)$, and $x_p$ to denote the point $F_{c}(m_p,n_p)$ within $lr$, 
\begin{equation}
\begin{aligned}
\mathcal{W}_{lsa}
&=W_{m_k,n_k}(m,n)\\
&=\frac{\exp(\theta(x)\cdot\phi(x_k))}{\sum\limits_{p\in lr}\exp(\theta(x)\cdot\phi(x_p))}
\end{aligned}
\label{eq:sim}
\end{equation}
where $\theta$ and $\phi$ are linear projection functions with learnable parameters for the query and key embeddings.
The local similarity aggregation (LSA) reduces the impact of outliers caused by lack of texture (\Cref{fig:LSA}).
We introduce the residual connection to the aggregation.
If $x$ denotes the center point $F_{2}(m,n)$, the LSA operation can be fully expressed as,
\begin{equation}
\begin{aligned}
x'&=x+\alpha\sum\limits_{k\in lr}\mathcal{W}_{lsa}\cdot\rho(x_k)
\end{aligned}
\end{equation}
where $\rho$ is a linear projection function with learnable parameters for the value embedding, making the aggregation operation match the traditional attention mechanism. $\alpha$ is a learnable parameter.

\subsection{Shifted Local Similarity Aggregation}
Aggregation between cost maps is also allowed. Because the elements within the cost map do not interact, the indices m and n are omitted for simplicity,
\begin{equation}
\begin{aligned}
C''(i,j)
&=\sum\limits_{(i_k,j_k)\in lr}\mathcal{W}_{slsa}\cdot\mathrm{Shift}(C'(i_k,j_k),rp)\\
&=\sum\limits_{(i_k,j_k)}^{lr}\big\langle \mathcal{W}_{slsa}\cdot F_1(i_k,j_k),\ \mathrm{Shift}(F_2,rp)\big\rangle
\end{aligned}
\end{equation}
where $lr$ represents the local region and $i_k$ and $j_k$ are the coordinates in the features $F_1$. In contrast to LSA, shifted local similarity aggregation (SLSA) has an additional shift operation to perform alignments of cost maps. $rp$ denotes the relative position of coordinate $(i_k,j_k)$ to the center point $(i,j)$. The cost maps are shifted based on this value (\Cref{fig:SLSA}).
We calculate the weights $\mathcal{W}_{slsa}$ by a formula similar to \cref{eq:sim}, 
except 
the parameters of the linear projection functions used in the SLSA are independent.
We use $x$ to denote the center point $F_{c}(i,j)$, $x_k$ to denote the weight point $F_{c}(i_k,j_k)$, and $x_p$ to denote the point $F_{c}(i_p,j_p)$ within $lr$, 

\begin{equation}
\begin{aligned}
\mathcal{W}_{slsa}
&=W_{i_k,j_k}(i,j)\\
&=\frac{\exp(\theta(x)\cdot\phi(x_k))}{\sum\limits_{p\in lr}\exp(\theta(x)\cdot\phi(x_p))}
\end{aligned}
\end{equation}
where $\theta$ and $\phi$ are linear projection functions with learnable parameters for the query and key embeddings.




\section{Experiments}
\noindent{\bfseries{Experimental Setup.}}
We follow the training schedule of previous work\cite{raft}.
We pre-train our model on FlyingChairs\cite{flownet} for 120k iterations with a batch size of 8 and FlyingThings3D\cite{things} for 160k iterations with a batch size of 6. 
We fine-tune the model on the combined dataset consisting of Sintel\cite{sintel}, KITTI-2015\cite{kitti}, and HD1K\cite{hd1k} for 160k iterations with a batch size of 6 and on KITTI-2015\cite{kitti} for 60k iterations with a batch size of 6.
We set the maximum learning rate to $2.5\times 10^{-4}$ in the FlyingChairs stage and $1.25\times 10^{-4}$ in the other stages.
Our model is built with the Pytorch\cite{pytorch} library and is trained on 2 GTX 2080Ti GPUs.

\begin{table*}[htb]
\caption{Quantitative results. We use brackets to represent the results on the training set.}
\label{tab:r1}
\centering
\begin{tabular}{@{}llc cc cc cc@{}}
\toprule
\multirow{2.5}{*}{Method} &\multirow{2.5}{*}{Year}
&\multicolumn{2}{c}{Sintel(train)}
&\multicolumn{2}{c}\qquad {KITTI-15(train)}
&\multicolumn{2}{c}{Sintel(test)}
&KITTI-15(test) \\
\cmidrule(r){3-4} \cmidrule(r){5-6} \cmidrule(r){7-8} \cmidrule(r){9-9}
&&Clean &Final
&Fl-epe &Fl-all
&Clean &Final
&Fl-all \\ 
\midrule

RAFT\cite{raft} &ECCV 2020 & (0.76)  & (1.22)  & (0.63)  & (1.5)  & 1.61  &2.86 &5.10\\
Flow1D\cite{flow1d}&ICCV 2021 & (0.84)  & (1.25)  & -  & (1.6)  &2.24  &3.81 &6.27\\
SCV\cite{scv}&CVPR 2021 & (0.86)  & (1.75)  & (0.75)  & (2.1)  &1.77  &3.88 &6.17\\
Separable Flow\cite{sflow}\qquad\qquad\qquad &ICCV 2021\qquad\qquad\quad &(0.69) & (1.10) &(0.69) &(1.60) &1.50 &2.67 &4.64\\
FlowFormer\cite{flowformer}&ECCV 2022 &(0.48) & (0.74)  &(0.53) &(1.11) &1.16 &2.09 &4.68\\
\midrule
LLA-Flow(ours) &- & (0.62)  & (1.00)  &  (0.57)  &  (1.23)  &1.50  & 2.62  & 5.01 \\
\bottomrule
\end{tabular}
\end{table*}

\noindent{\bfseries{Quantitative Results.}}
Our model outperforms RAFT on the Sintel final pass (\Cref{tab:r1}), with a reduction in AEPE from 2.86 to 2.62. Compared to Sintel clean pass, the Sintel final pass adds additional motion blur and atmospheric effect, making the optical flow estimation more challenging. FlowFormer inserts the GMA module\cite{gma} for occlusion processing and uses the ImageNet-pretrained transformer, achieving state-of-the-art performance on standard benchmarks.

\noindent{\bfseries{Visual Results on Textureless Regions.}}
We visualize challenging examples from the Sintel final pass (\Cref{fig:test}). The metric results are annotated in the lower left corner. EPE represents average endpoint error and S40+ represents endpoint error over regions with velocities larger than 40 pixels. Regions with high velocities usually are strongly influenced by motion blur and lack of texture. Our approach outperforms RAFT and Separable Flow on S40+.
The visual results show that our method performs better than RAFT and Separable Flow and is comparable to Flowformer in the textureless regions (marked by the black dashed boxes), which demonstrates the effectiveness of our modules.

\begin{table}[htb]
\caption{Ablation results. Training on Chairs and Things3D.}
\label{tab:ablation}
\centering
\begin{tabular}{@{}ll cc c @{}}
\toprule
\multirow{2.5}{*}{Experiment}
&\multirow{2.5}{*}{Method} 
&\multicolumn{2}{c}{Sintel(train)}
&\multirow{2.5}{*}{Parameters} \\
\cmidrule(r){3-4}
&&Clean &Final\\ 
\midrule

\multirow{4}{*}{Local Agg.}
& No    &1.48   &2.68     &5.25M \\
& Only LSA    &1.43   &2.65    &5.35M \\
& Only SLSA    & 1.41   &2.64   &5.42M \\
& \underline{All}  &{\bfseries 1.40}   &{\bfseries 2.61}  &5.52M \\
\midrule

\multirow{2}{*}{Shift Operation}
& No Shift    &1.42   &2.70     &5.45M \\
& \underline{Shift}  &{\bfseries 1.40}   &{\bfseries 2.61}    &5.52M \\
\midrule

\multirow{3}{*}{Local Region}
& 3x3    &{\bfseries 1.38}   &2.85     &5.52M \\
& \underline{5x5}    & 1.40   &{\bfseries 2.61}     &5.52M \\
& 7x7    &1.46   &2.73     &5.52M \\
\midrule

\multirow{2}{*}{Position Enc.}
& \underline{No}   &1.40   &{\bfseries 2.61}     &5.52M \\
& Sinusoidal &{\bfseries 1.35}   &2.75    &5.52M \\
\midrule

\multirow{2}{*}{Context Fusion}
& \underline{No}    &{\bfseries 1.40}   &{\bfseries 2.61}     &5.52M \\
& Yes &1.43   &{2.67}     &5.61M \\

\bottomrule
\end{tabular}
\end{table}

\noindent{\bfseries{Ablation Study.}}
We evaluate the model performance under various settings (\Cref{tab:ablation}).
First, the model’s capacity to handle the lack of texture weakens in the absence of local aggregation, resulting in decreased performance.
We evaluate the models with the LSA module only or the SLSA module only, respectively, and observe that both show performance improvements over using no aggregation.
Moreover, we remove the shift operation from the SLSA module so that the cost maps of surrounding pixels are not aligned and are summed directly.
In practice, this modification is identical to aggregating the first frame with the LSA module.
The results show that the additional shift operation yields performance improvements.
The size of the region for the local aggregation module is important.
Applying too large local aggregation regions may degrade the performance. The distant pixels may provide some false constraints because they are not closely associated with the center pixel. 

We add sinusoidal position encoding on the context features to learn the characteristics of datasets.
While this strategy slightly improves the performance of our method on the clean pass, it performs poorly on the final pass that contains more textureless regions.
Furthermore, we attempt to extract the context features from both frames to provide more comprehensive context information.
Specifically, we modify the input of the context encoder to be the paired frames, allowing it to extract context features $F_{c1}$ and $F_{c2}$ from them. The model utilizes these context features separately to guide the SLSA and LSA modules.
Then we use the fused context features $\tanh(F_{c_1}-F_{c_2})$ to initialize the GRU hidden state.
The ablation demonstrates that additional context fusion slightly weakens the performance of the model.
We speculate that it is caused by the lack of alignment between the two features.

\noindent{\bfseries{Parameter, Memory and Runtime.}}
We thoroughly examine the costs of different aggregation methods, including the number of parameters, training memory, inference memory, and runtime (\Cref{tab:runs}).
We complete our test experiments on a single GTX 2080Ti GPU and report the average of 20 runs.
For training, the resolution of frames is $368\times 496$ and we record the results for a batch size of 1 and a batch size of 4. We set the number of GRU iterations to 12.
For inference, The resolution of frames is $440\times 1024$ with a batch size of 2 and we set the number of GRU iterations to 32.
The aggregation network in Separable Flow applies for C++/CUDA extensions. 
Because no detailed data are provided in the original paper, we present our reproduction results.

Compared to RAFT, our method only increases the parameters by 0.26M, which is much lower than the global aggregation methods.
FlowFormer achieves state-of-the-art performance but requires the most extra memory for both training and inference.
Furthermore, our method requires much less runtime than other methods in the \Cref{tab:runs}, while maintaining competitive performance in the textureless regions.

\begin{table}[t]
\centering
\caption{Parameter, memory and runtime statistics. Trn-M(1/4) = Training memory(bs=1/bs=4). Inf-M = Inference memory. Time is inference time or runtime. OOM = Out of memory. S-Flow means the Separable Flow.}
\label{tab:runs}
\begin{tabular}{@{}lcccc@{}}
\toprule
Method   & Parameter   &Trn-M(1/4)        & Inf-M     &Time \\ \midrule
RAFT       & 5.26M    & 2.4G/5.8G          & 0.27G            & 0.27s          \\
S-Flow     & 8.35M    & 3.5G/9.8G          & 0.29G            & 1.31s          \\
FlowFormer & 16.2M    & 9.8G/OOM           & 0.97G            & 1.02s          \\ 
Ours       & 5.52M    & 2.9G/7.5G          & 0.39G            & 0.34s          \\ \bottomrule
\end{tabular}
\end{table}

\vspace{8pt}
\section{Conclusion}
The lack of texture challenge often occurs in the optical flow estimation and leads to ambiguities in cost volumes.
In this paper, we design the local similarity aggregation and the shifted local similarity aggregation modules for cost volumes to reduce the impact of the outliers. 
The aggregation for the cost volume is implemented with lightweight modules that act on the feature maps. 
We comprehensively compare the differences in parameter, memory, and runtime between our method and the methods that perform global aggregation on cost volumes.
The visual results and running statistics demonstrate that our method maintains competitive performance at a lower cost.

\clearpage
\clearpage
\bibliographystyle{IEEEbib}
\bibliography{strings,refs}

\end{document}